\documentclass{article}

\usepackage{arxiv}

\usepackage[utf8]{inputenc} 
\usepackage[T1]{fontenc}    
\usepackage{hyperref}       
\usepackage{url}            
\usepackage{booktabs}       
\usepackage{amsfonts}       
\usepackage{nicefrac}       
\usepackage{microtype}      
\usepackage{lipsum}		
\usepackage{graphicx}
\usepackage{natbib}
\usepackage{doi}

\title{\emph{negativas}: a prototype for searching and\\ classifying sentential negation in speech data}

\author{ \href{https://orcid.org/0009-0000-5270-8033}{\includegraphics[scale=0.06]{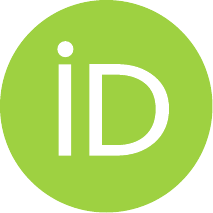}\hspace{1mm}Túlio Sousa de Gois} \\
	Laboratório Multiusuário \\
	de Informática e Documentação\\
    Linguística\\
    Federal University of Sergipe\\
	\texttt{tuliosg@academico.ufs.br} \\
	\And
    \href{https://orcid.org/0000-0002-2244-8960}{\includegraphics[scale=0.06]{orcid.pdf}\hspace{1mm}Paloma Batista Cardoso} \\
	Laboratório Multiusuário \\
	de Informática e Documentação\\
    Linguística\\
    Federal University of Sergipe\\
	\texttt{paloma-batistacardoso@hotmail.com} \\
}

\date{}

\renewcommand{\headeright}{}
\renewcommand{\undertitle}{}
\renewcommand{\shorttitle}{\emph{negativas}: a prototype for searching and classifying sentential negation in speech data}

\begin{document}
\maketitle

\begin{abstract}
Negation is a universal feature of natural languages. In Brazilian Portuguese, the most commonly used negation particle is não, which can scope over nouns or verbs. When it scopes over a verb, não can occur in three positions: pre-verbal (NEG1), double negation (NEG2), or post-verbal (NEG3), e.g., não gosto, não gosto não, gosto não ("I do not like it"). From a variationist perspective, these structures are different forms of expressing negation. Pragmatically, they serve distinct communicative functions, such as politeness and modal evaluation. Despite their grammatical acceptability, these forms differ in frequency. NEG1 dominates across Brazilian regions, while NEG2 and NEG3 appear more rarely, suggesting its use is contextually restricted. This low-frequency challenges research, often resulting in subjective, non-generalizable interpretations of verbal negation with não. To address this, we developed \emph{negativas}, a tool for automatically identifying NEG1, NEG2, and NEG3 in transcribed data. The tool's development involved four stages: i) analyzing a dataset of 22 interviews from the Falares Sergipanos database, annotated by three linguists, ii) creating a code using natural language processing (NLP) techniques, iii) running the tool, iv) evaluating accuracy. Inter-annotator consistency, measured using Fleiss’ Kappa, was moderate (0.57). The tool identified 3,338 instances of não, classifying 2,085 as NEG1, NEG2, or NEG3, achieving a 93\% success rate. However, \emph{negativas} has limitations. NEG1 accounted for 91.5\% of identified structures, while NEG2 and NEG3 represented 7.2\% and 1.2\%, respectively. The tool struggled with NEG2, sometimes misclassifying instances as overlapping structures (NEG1/NEG2/NEG3). These challenges stem from the dataset's lack of punctuation, which in written texts, marks sentence boundaries. In spoken data, prosodic cues serve this purpose, recognized by speakers but not by the tool. This highlights the need for advancements in NLP to better handle the unique features of spoken language data.
\end{abstract}


\section{Introduction}
Negation is a common phenomenon in all natural languages \citep{dahl2010typology} and has been the focus of numerous descriptive linguistics studies, from different perspectives \citep{rocha2013negaccao, goldnadel2016funccoes, oliveira2022variaccao}. In Brazilian Portuguese (BP), negative structures can be formed by \textit{não} (no/not) in pre-verbal, double, and post-verbal positions (NEG1, NEG2, NEG3, respectively), as exemplified in (1): 

\begin{itemize}
\item[(i)]  \begin{quote}
(Eu) Não gosto (NEG1) \newline
\textit{(I) don’t like it}
\end{quote}

\item[(ii)] \begin{quote}
(Eu) não gosto não (NEG2)\newline
\textit{(I) don’t  like it  no}
\end{quote}

\item[(iii)] \begin{quote}
(Eu) gosto não (NEG3)\newline
\textit{(I don’t) like it no}
\end{quote}
\end{itemize}

From a variationist perspective, these three negative structures are three forms to express opposition. From a pragmatic perspective, NEG1, NEG2, and NEG3 assume different communicative functions. These two approaches to negation with \textit{não} in BP are commonly rooted in the analysis of sociolinguistic interviews. 

These interviews are traditionally structured as dialogues between the informant and the documenter, lasting an average of 50 minutes. During this time, a large volume of speech data is generated. Without the use of automatic tools to search for a specific element of interest, handling databases made up of dozens or hundreds of sociolinguistic interviews is a challenge. This is why, possibly, several studies about negative structures with \textit{não} are carried out with a reduced sample of analysis. Consequently, many conclusions about the use of NEG1, NEG2, and NEG3 are based on subjective interpretations that cannot be generalized, which could be made through the use of sophisticated statistical analysis like models of conditional inference three \citep{freitag2020modelo}. But, for this, is it necessary to have a large database of the phenomenon under analysis. 

The lack of studies about the use of NEG1, NEG2, and NEG3 executed with large samples is also prejudicial to the development of technologies based on the recognition of language patterns. It is impossible to develop efficient Artificial Intelligence (AI) models to process negative structures with \textit{não} without a reliable description. To contribute to studies about negation in BP and the processing of natural language, we present the \emph{negativas} tool, which aims to make it possible to search, automatically, for negation structures formed with the adverb \textit{não} in pre-verbal, double, and post-verbal positions. The tool was built using the Python language and libraries for NLP and data science.

\section{Negative structures with \textit{não} in Brazilian Portuguese (BP)}

Negation is a property that can be expressed in different ways: by (i) morphemes or affixes (\textit{infeliz, desiludido} – unhappy, disillusioned); (ii) negative particles (\textit{não, nunca} – no, never); and (iii) negative verbs (\textit{inviabilizar, impedir} – make it unviable, to stop).

In BP, one of the negative particles that can indicate negation is \textit{não}. Traditionally, negation is understood as a strategy to indicate opposition to affirmation. From this perspective, NEG1, NEG2, and NEG3 have the same function and are, therefore, variants of the same variable. The approach of negation as opposition is heavily influenced by studies in logic. From this perspective, the propositional content of sentences is analyzed based on truth values, divided into true or false.

In the field of linguistic studies, numerous studies have focused on the description and analysis of negative structures with \textit{não}. Variationist approaches assume that negative structures with \textit{não} in pre-verbal, double, and post-verbal positions are three variants of the same variable. From this point of view, the use of NEG1, NEG2, and NEG3 are conditioned by linguistic (type of sentence in which \textit{não} occurs, presence/absence of subject, type of verb, presence/absence of other negative adverbs) and social factors (level of education, age group, local of residence) \citep{rocha2013negaccao, oliveira2022variaccao}, suggesting that these uses may be a factor of regional identification since the occurrence of NEG3 is more present among speakers from the Northeast of Brazil.

Another approach to describing and analyzing negative structures with \textit{não} in BP assumes that NEG1, NEG2, and NEG3 are not variants of the same variable. From this perspective, the three structures have specific pragmatic functions. Speakers can use a negative structure with \textit{não} to indicate opposition and they can also use it to appease an information, to be polite to their listeners:
\begin{quote}
\begin{itemize}
\item[(1)] 
DOCLS: mas como os colegas do teu curso como que é a relação?\newline
\textit{But with your classmates, how is your relationship?}

GUI1MI: (...) assim eu go/ gosto não tenho problema com grande parte mas tem uns que assim não é que eu não tenha problema evito né (...)\newline
\textit{I mean, I li/ like, I don’t have problems with most of them, but there are some of them that is not that I don’t have problems with them, I even avoid right (...)}
\end{itemize}
\end{quote}

From a pragmatic point of view, negation has interactional properties and is therefore characterized as an act performed by the speaker in front of the listener. Approaches of this type also have a large place in studies describing Brazilian Portuguese \citep{goldnadel2016funccoes, Goldnadel_Petry_Lamberti_2020}, especially those derived from \citet{schwenter2005pragmatics} assumptions. From this perspective, the uses of NEG1, NEG2, and NEG3 are conditioned by the informational status of the negated content, which can be new or old, expressed implicitly or explicitly.

To execute descriptive studies that can test different theories about the uses and functions assumed by NEG1, NEG2, and NEG3, it is necessary to make analyses based on a large volume of data, which is impossible to do by searching, manually, for each occurrence of negative structures with \textit{não} in a set of sociolinguistic interviews. One possibility to deal with this problem is to use spaCy, a library for Natural Language Processing (NLP) which has models to recognize syntactic patterns of Brazilian Portuguese.

\section{spaCy}

spaCy \citep{honnibal2020spacy} is an open-source library for advanced natural language processing in Python. Its features and functionalities are related to linguistics, such as tokenization and data labeling, and also to machine learning (ML), such as model training.  In addition to the variety of resources applied to NLP, spaCy also stands out for its pre-trained models for different languages, with 80 pipelines trained for 24 languages. For Brazilian Portuguese (BP), this library has 3 pre-trained models, differing in their size, training datasets, and functionalities. In this work, we use pt\_core\_news\_lg\footnote{https://spacy.io/models/pt\#pt\_core\_news\_lg}, a pipeline recommended for tasks in which greater accuracy is required. To \emph{negativas}, three spaCy functionalities are central to its development and operation: tokenization, POS-tagging, and Rule-based Matching.

Tokenization is the process of segmenting texts into tokens (words, punctuation, etc.). In spaCy, this is a non-destructive process because, after segmentation, the tokens can be used to reconstruct the original sentence without any loss. This task is carried out according to the syntax of the language being processed. So, when we load the pre-trained model for BP, we are configuring specific syntactic specifications for processing the texts.

Through pre-trained pipelines, spaCy can predict the attributes of each token, depending on its context. These include part-of-speech tagging, which assigns a tag corresponding to the grammatical class of each word. The tags adopted by the library are the Universal POS tags\footnote{https://universaldependencies.org/u/pos/\#universal-pos-tags} \citep{rademaker2017universal}. 

The attributes of the tokens not only make it possible to understand more about the word and the syntactic context in which it is inserted but also help with tasks such as searching for certain phenomena in a large corpus. This process is carried out by assigning patterns to rule-based matching. Thus, it is possible to create search expressions using the grammatical class labels of the tokens (among other attributes, such as the text itself) and insert customized nomenclatures for each pattern created. This is how the search for structures with \textit{não} is performed by \emph{negativas}.

\section{Methodology}

The process of developing and validating \emph{negativas} took place in four main stages: i) understanding the sample Deslocamentos 2020 dataset, formed by sociolinguistic interview 22 interviews, and the NEG1, NEG2, and NEG3 structures; ii) building the code based on NLP techniques; iii) running the tool, and iv) analyzing accuracy if the results achieved.

\subsection{The sample Deslocamentos 2020}

The Deslocamentos 2020 sample is part of the Falares Sergipanos database \citep{freitag2013banco}. In total, this sample consists of 100 audio-recorded sociolinguistic interviews, structured in dialogues between informants and documenters. The informants are undergraduate students at the Federal University of Sergipe, and the documenters are researchers linked to the Grupo de Estudos em Linguagem, Interação e Sociedade – GELINS.

All the interviews that compose the Deslocamentos 2020 sample were audio-recorded and then saved in \textit{.wav} format. They were then transcribed using ELAN software \citep{brugman2004annotating}. The interviewee's and interviewer's speeches were recorded on individual tracks. Separately, on a third track, speech disfluencies, such as pauses and hesitations, were recorded. This procedure is justified by the possibility of generating .txt files containing the individual content of each track at a later date. In these files, we included a header containing information about the place where the interview was carried out, the speaker's gender and age, city of origin and residence, and period of undergraduate study. All the text files were used to create a single dataset with data from 22 interviews\footnote{The Deslocamentos 2020 sample consists of 100 sociolinguistic interviews. Of these, 32 were audio and video recorded. The construction of the protocol using negatives was motivated by a Doctoral project whose goal is to describe the functions of negative structures with “não”, considering, besides linguistic structure, and and face gestures.  More information about it is presented by \citet{batista2023speech} and \citet{cardoso2025entre}. } , which was then used for the search task with the proposed tool.

\section{\emph{negativas}}

The tool \emph{negativas}\footnote{All code is available on OSF through the following link:\url{https://osf.io/3v4u9/}} was developed using Colab, a Google service that allows Jupyter notebooks to be created. Besides that, we used \textit{spaCy} libraries to apply NLP techniques, while pandas were used to structure and save the data.

In the dataset, each file has a header to identify information about the interview. Because of the transcription standards adopted by GELINS, the transcription files also have pauses and noise markings throughout the text (speech disfluences). Therefore, the data pre-processing tasks involved extracting the header information and cleaning the data by removing the aforementioned markings.

The search and classification of pre-verbal, double, and post-verbal negation structures started with the use of Matcher, from the spaCy library, a rule-based matching mechanism that allows searches to be carried out using token attributes as parameters. The search patterns are lists of dictionaries, and each dictionary represents a token. By adding them to the Matcher, it is possible to enter an ID (identifier), a structure used to classify each pattern found. For the phenomenon studied, we searched for forms corresponding to the negation structures presented (NEG1, NEG2, and NEG3). The patterns created used exact agreement with the word "\textit{não}" and POS tags to identify verbs and auxiliaries as attributes, and finally, corresponding IDs were inserted for each structure. The attributes used as a basis for building the patterns, as well as their IDs, are shown in Table 1.

\begin{table}
\begin{center}
\begin{tabular}{|p{3cm} | p{3cm} | p{9cm}|}
    \hline
    \textbf{Negation structure} & \textbf{ID} & \textbf{Basic attributes} \\
    \hline
    NEG1 & pre-verbal & [\{"TEXT": "não"\}, \{"POS": \{"IN": ["VERB", "AUX"]\}\}] \\
    \hline  
    NEG2 & double negation & [\{"TEXT": "não"\}, \{"POS": \{"IN": ["VERB", "AUX"]\}\}, \{"TEXT": "não"\}]\\
    \hline  
    NEG3 & post-verbal negation & [\{"POS": \{"IN": ["VERB", "AUX"]\}\}, \{"TEXT": "não"\}]\\
    \hline
\end{tabular}
\end{center}
\caption{Search pattern attributes}
\end{table}

The occurrences found and their classifications are stored in spreadsheets, which
contain the interview header data, the type of negation found and its identifier, in this case, the negation structure corresponding to the occurrence (NEG1, NEG2, NEG3).

\subsection{Classification metrics}

In order to validate the classification performed by \emph{negativas}, we generated a confusion matrix and calculated the main metrics based on the annotation of the data by three linguists. This stage was divided into two parts: the calculation of agreement between annotators and the evaluation of the tool's metrics.

Calculating agreement between annotators is necessary to assess the consistency of data annotations. In this work, the measure used was Fleiss' Kappa, which can be used with more than two annotators. It was applied using the \textit{multi\_kappa function} from the NLTK library \citep{loper2002nltk}. In addition, Cohen's Kappa was used to assess the agreement between the annotators, using the \textit{cohen\_kappa\_score} function from the \textit{sci-kit learn library} \citep{pedregosa2011scikit}. After calculating the concordance, we unified the annotations by calculating the frequency of occurrence of negative structures. For each piece of annotated data, its final label was the one with the highest number of appearances in the annotations.

The classification was evaluated using the function \textit{confusion\_matrix} and \textit{classification\_report} also from sci-kit learn, to generate the confusion matrix and calculate precision, recall, F1 and accuracy, respectively. The Cohen's kappa coefficient was also calculated between human annotations and the classifications to determine the level of concordance.

\section{Results}

\begin{figure}[!ht]
    \centering
    \includegraphics[width=0.5\linewidth]{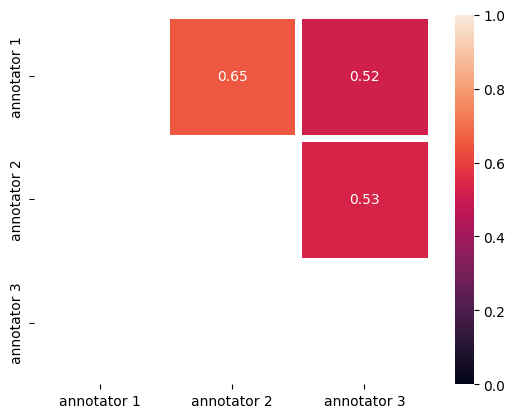}
    \caption{Inter-annotation agreement}
    \label{fig:inter-annotator}
\end{figure}

The data annotation obtained 0.57 of agreement (Fleiss' Kappa), which is considered a moderate value. Cohen's Kappa was calculated for each pair of annotators and generated values between 0.52 and 0.65 (Figure \ref{fig:inter-annotator}). Human annotations resulted in 1893 (90.8\%) occurrences of the pre-verbal structure, 92 (4.4\%) occurrences of the post-verbal structure, and 100 (4.8\%) of double negation.

\newpage

The application of \emph{negativas} to the dataset resulted in the identification of 3338 occurrences of the word \textit{não}, of which 2085 were classified as one of the three sentential negative structures.
Pre-verbal negative structures (NEG1) were predominant, accounting for 91.5\% of the recognized structures. Following this order, NEG3 (post-verbal) was identified in 7.2\%, and finally, NEG2 (double negation), which occurred in only 1.2\% of the negations.

\begin{figure}[!ht]
    \centering
    \includegraphics[width=0.7\linewidth]{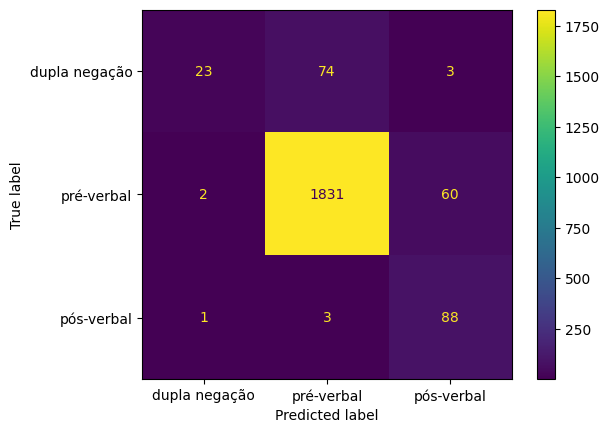}
    \caption{negativas confusion matrix}
    \label{fig:confusion-matrix}
\end{figure}

At the stage of classifying negation structures, the \emph{negativas} showed an accuracy of 93\%. Besides this global metric, we also calculated precision, revocation, and F1 to  each negation structure, as can be seen in Table 2. We calculated all the metrics from the elements present in the confusion matrix (see Figure \ref{fig:confusion-matrix}).

The agreement between human annotations and \textit{negativas} classifications was calculated using Cohen's Kappa coefficient, yielding moderate agreement ($\kappa = 0.58$).

\section{Discussion}

Linguistic analyses based on speech data deal with a large volume of data. To deal with this amount of information, it is necessary to develop tools that make it possible to automate searches for the subsequent analysis of specific phenomena.

In the field of descriptive linguistics, studies are traditionally based on manual data handling. Transcription tools such as ELAN make it possible to search for specific items but in a generalized way. Searching for occurrences of the adverb "\textit{não}" results in all the contexts in which this word is used. Once you're interested in analyzing specific usage situations - such as \textit{não} in pre-verbal, double, and post-verbal positions – filtering out specific syntactic structures becomes a long and expensive job. This is why the use of automatic tools is a latent need.

The different possibilities of negation with \textit{não} do not occur randomly: they might be conditioned by linguistic, social, and pragmatic factors. Understanding these factors is important to describe how this universal property of natural languages works in BP. Understanding them also contributes to the development of efficient machine learning and AI models that adequately produce and process NEG1, NEG2 and NEG3. To do this, it is essential to work with a large volume of data, which is only feasible through the use of automatic tools whose efficiency is continually improved.

\emph{negativas} has proved to be efficient in finding and classifying negative structures with \textit{não}, but it does have its limitations. Such as the occurrence of double negation, which in some cases is not recognized as an individual syntactic structure, returning an occurrence that is simultaneously classified as NEG1, NEG2, and NEG3, and also the treatment of intervening material present in the data, which ends up making it difficult to model the search patterns. Another notable limitation relates to the data imbalance, with NEG1 structures being significantly more frequent than other categories, which adversely affects precision when calculating classification metrics.

The tool's difficulty in recognizing the boundaries between NEG1, NEG2 and NEG3 is due to the type of data submitted. The interview transcripts in the Falares Sergipanos database have no punctuation marks, except the question mark. Elements which, in written data, delimit sentence boundaries and constituents (period, comma), are not present in the Displacements 2020 sample. In speech, these boundaries are delimited by prosodic parameters. Native speakers, due to the prosodic structure of Portuguese, perceive \textit{não} fui \textit{não} (I didn’t went no) as a single block. The machine does not. This fact indicates the limitations of the tool and also the need for improvements in natural language processing techniques so that they take into account the particularities of data submitted to automatic tools.

\section{Conclusions}
Given that \emph{negativas} use a pre-trained model, the metrics calculated represent the efficiency of the patterns built using the pipeline in identifying negation structures. While the tool achieved a high accuracy of 93\%, this metric primarily reflects the data imbalance rather than true performance. However, the Cohen's Kappa coefficient ($\kappa = 0.58$) between \textit{negativas} and human classifications indicates moderate agreement, demonstrating the tool's potential to automate the description of a large volume of data on negative structures with \textit{não} in pre-verbal, double and post-verbal positions. However, to do this, it is necessary to overcome the limitations of speech data processing. Furthermore, the patterns built for the classification stage carried out by the tool for speech data also allow for the analysis of negation data from writing. However, for data from social networks, for example, changes to the structure would be necessary to accommodate different variations of \textit{não} such as "n" and "ñ".

Finally, the process of building \emph{negativas} has shown that natural language processing models need to take into account the origin and characteristics of the data to be effectively efficient, thus contributing to advances in the fields of linguistics and AI.

\bibliographystyle{unsrtnat}
\bibliography{references}

\end{document}